\documentclass[letterpaper, 10 pt, conference]{ieeeconf}  % Comment this line out if you need a4paper

\IEEEoverridecommandlockouts                              % This command is only needed if 
\usepackage{amsmath} % assumes amsmath package installed
\usepackage{amssymb}  % assumes amsmath package installed
\usepackage{algorithm}  % For algorithm environments
\usepackage{algorithmic}  % For algorithm content
\usepackage{listings}
\usepackage{xcolor}
\usepackage{booktabs}
\usepackage{graphicx}    % For including images
\usepackage{subcaption}  % For subfigures with individual captions
\usepackage{cite}

\title{\LARGE \bf
Look Before You Leap: Using Serialized State Machine \\for Language Conditioned Robotic Manipulation 
}

\author{Tong Mu$^{1}$, Yihao Liu$^{1,2,*}$, Mehran Armand$^{1,2}$ % <-this % stops a space
\thanks{This work was supported by National Institute of Arthritis and Musculoskeletal and Skin Diseases (R01AR080315) and National Institute of Biomedical Imaging and Bioengineering (R01EB023939). }
\thanks{$^{1}$ Department of Computer Science, Johns Hopkins University, Baltimore, MD 21218, USA}%
\thanks{$^{2}$ The Institute for Integrative and Innovative Research, University of Arkansas, Fayetteville, AR 72701, USA}
\thanks{* Corresponding author {\tt\small (yliu333@jhu.edu)}}
}

\begin{document}

\maketitle
\thispagestyle{empty}
\pagestyle{empty}

%%%%%%%%%%%%%%%%%%%%%%%%%%%%%%%%%%%%%%%%%%%%%%%%%%%%%%%%%%%%%%%%%%%%%%%%%%%%%%%%
\begin{abstract}

Imitation learning frameworks for robotic manipulation have drawn attention in the recent development of language model grounded robotics. However, the success of the frameworks largely depends on the coverage of the demonstration cases: When the demonstration set does not include examples of how to act in all possible situations, the action may fail and can result in cascading errors. To solve this problem, we propose a framework that uses serialized Finite State Machine (FSM) to generate demonstrations and improve the success rate in manipulation tasks requiring a long sequence of precise interactions. To validate its effectiveness, we use environmentally evolving and long-horizon puzzles that require long sequential actions. Experimental results show that our approach achieves a success rate of up to 98\% in these tasks, compared to the controlled condition using existing approaches, which only had a success rate of up to 60\%, and, in some tasks, almost failed completely.

\end{abstract}

%%%%%%%%%%%%%%%%%%%%%%%%%%%%%%%%%%%%%%%%%%%%%%%%%%%%%%%%%%%%%%%%%%%%%%%%%%%%%%%%
\section{Introduction}

The recent advancements in combining Large Language Models (LLMs) with robotic systems have shown potential in automating complex task planning and execution \cite{llm1, llm2, gensim, hua2024gensim2, wu2023tidybot, hoeg2022more, liang2023code}. Yet, LLMs have limited capacity for translating high-level task description text into strong, executable policies over manipulation tasks with long horizons. Existing approaches, such as those that leverage LLMs to generate task demonstration code for imitation learning \cite{gensim, hua2024gensim2}, extend on the classic philosophy of using language-conditioned policies to set goals and transfer concepts across different tasks \cite{cliport}. The required training after the demonstration collection makes the framework data-dependent. These demonstrations can be human-curated \cite{cliport} or LLM-generated \cite{gensim, hua2024gensim2}, and they often rely on an assumption for the task where all the objects are randomly spawned in simulation when generating a demonstration. 

While the current approaches work well for short-horizon tasks where only a few sequential actions are required or loosely constrained tasks where precise positioning or ordering is less critical, they may not scale to scenarios that require precise state-dependent reasoning. For example, the reliance on vision for imitation learning \cite{cliport, gensim} leads to failures in manipulation when encountering previously unseen environmental structures during the execution of long-horizon tasks. 

\begin{figure}[t]
    \centering
    \begin{subfigure}[b]{0.48\columnwidth}
        \includegraphics[width=\textwidth]{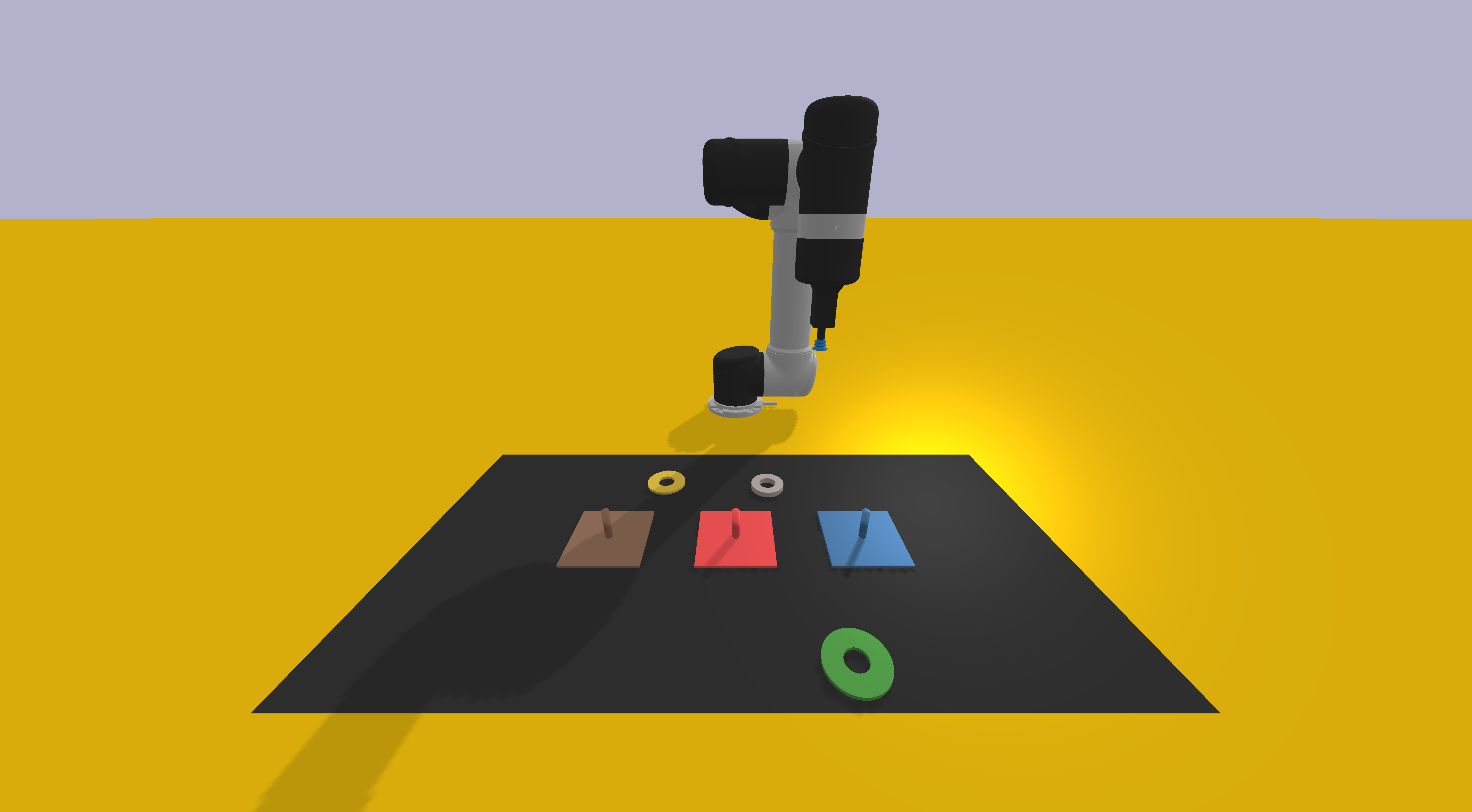}
        \caption{The scene with all randomly spawned objects.}
        \label{fig:hanoirand}
    \end{subfigure}
    \hfill
    \begin{subfigure}[b]{0.48\columnwidth}
        \includegraphics[width=\textwidth]{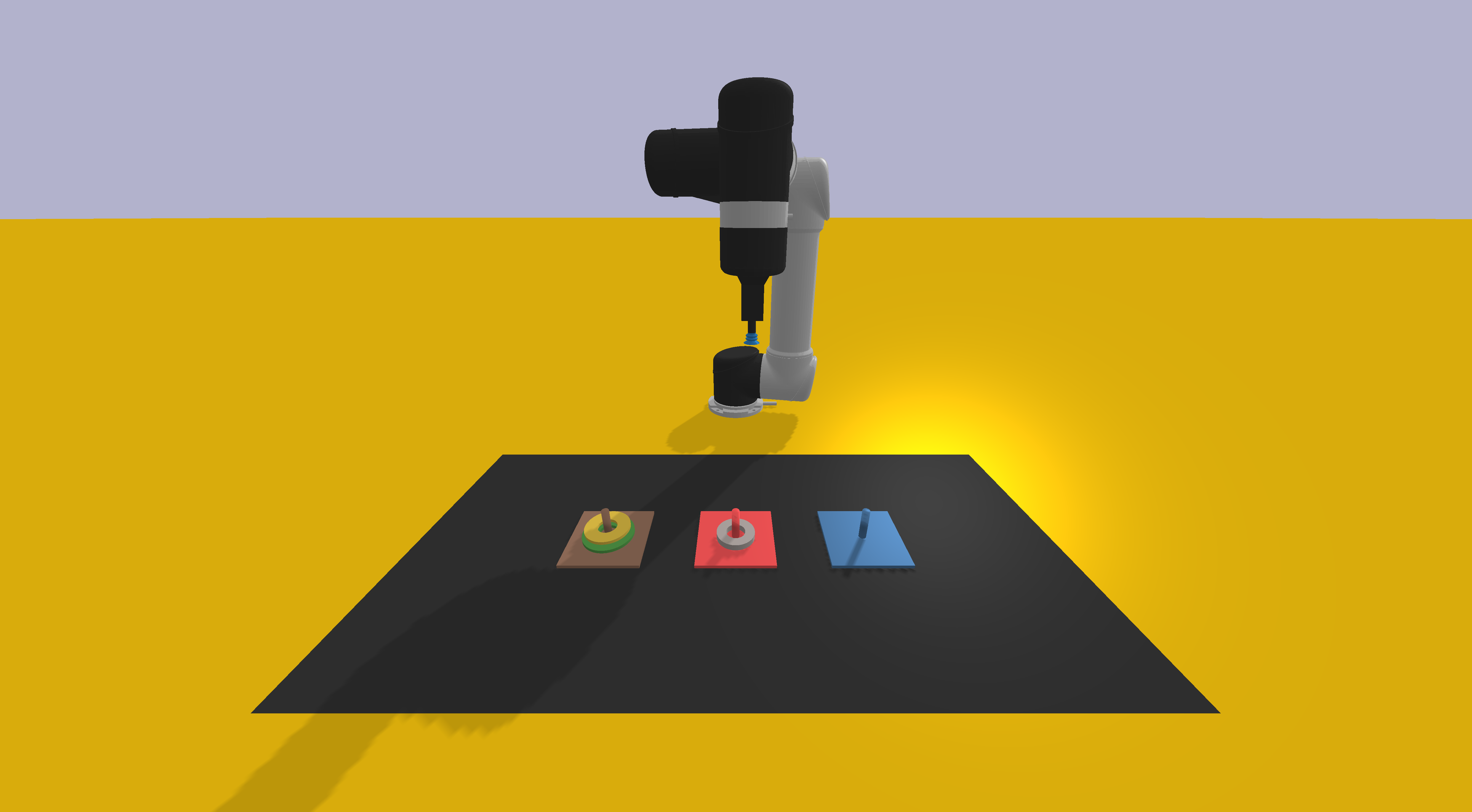}
        \caption{The actual scene of a state when solving Towers of Hanoi.}
        \label{fig:hanoiactual}
    \end{subfigure}
    \caption{Comparison between a randomly initialized scene using the existing approach for data collection (a) and the actual scene the system encounters when completing the Towers of Hanoi task (b).}
    \label{fig:compare}
\end{figure}

This limitation becomes evident in our experimental findings: Current frameworks are prone to cascading failure in these tasks using vision. For example, in Tower of Hanoi \cite{cliport, gensim}, the primitive tasks are to move a ring in a certain color to a stand in another color, as shown in Fig. \ref{fig:compare}. Existing approaches with objects spawned arbitrarily on the table are trained on randomly initialized sub-task demonstrations. They may place rings on incorrect stands or miss valid pick points during the tests. This occurs because randomized training distributions fail to capture the structural consistency required for sequential spatial reasoning (e.g., stand/ring relationships depend on prior moves). In other words, when objects are subject to dynamically evolving spatial constraints, policies trained on randomly initialized demonstrations can experience performance degradation due to the divergence of the training and the evolving execution environment structures. 

\begin{figure*}[t]
    \centering
    \includegraphics[width=\textwidth]{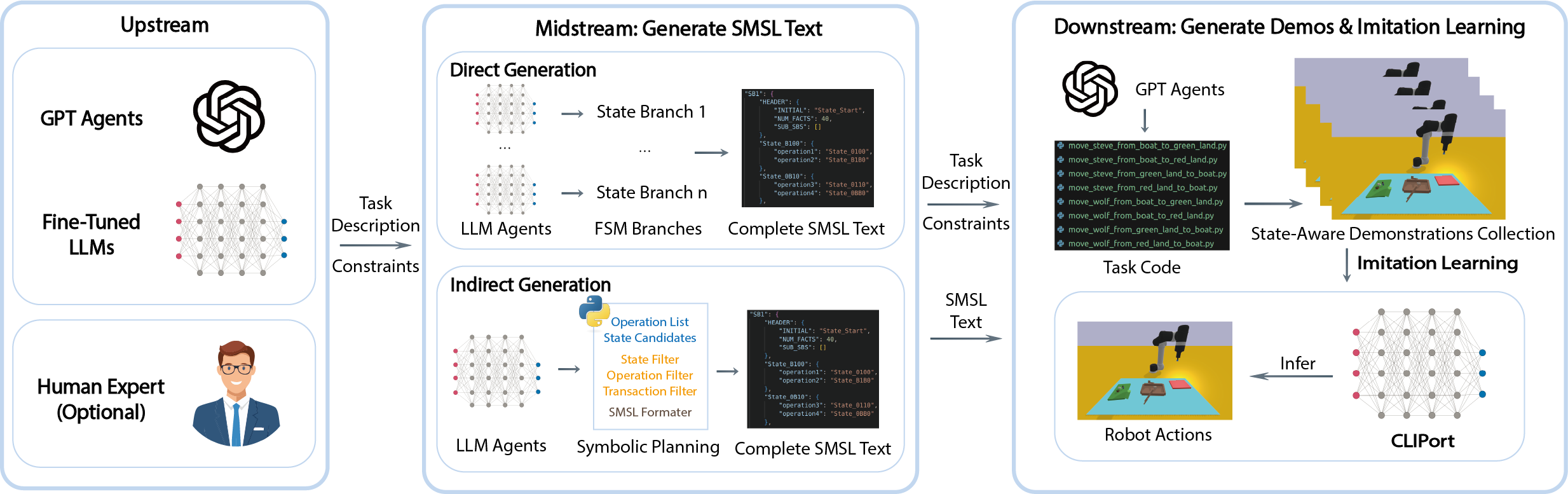}
    \caption{Overall architecture of the proposed method. The upstream processes the high-level task goal to provide detailed task descriptions with constraints to the midstream. We propose two methods in the midstream: The direct generation is to split the whole SMSL generation into multiple state branches to be generated separately and then integrate them. Indirect generation will use LLM to generate a symbolic planning script to plan and format an SMSL text. The LLM agents in downstream with engineered prompts will take this SMSL text along with the task description and constraints from the upstream to generate the task demonstration code. We will then use the task code to collect the dataset for imitation learning in a ``state-awared'' workflow.}
    \label{fig:architecture}
\end{figure*}

Our work aims to address this issue by dividing complex tasks into sub-actions, predicting possible intermediate environment structures, and training the imitation framework on all or most predicted states. These can be concluded into two key principles: ``\textit{divide and concur}'' and ``\textit{look before you leap}''. To achieve this, we propose a state-aware imitation learning pipeline that systematically links finite state machine (FSM) with environment-conditioned robot policy training (Fig. \ref{fig:architecture}). We validate our framework on long-horizon tasks and found that it improves the success rate from up to 60\% in existing approaches to almost 100\%. In summary, the proposed framework has the following features that distinguish this work from previous research:

\begin{itemize}
    \item By introducing the State Machine Serialization Language (SMSL) as a guide into demonstration generation for imitation learning, the system ensures that the resulting policy training is aware of the precise geometric constraints of complex long-horizon tasks.
    \item The planning process is highly deterministic because our framework uses SMSL’s state transitions to propagate and record the geometric configuration of objects at each step of the task. These configurations are stored as persistent environmental states, enabling deterministic scene initialization for demonstrations.
    \item The predicted states are filtered with constraints to ensure the correctness, and the entire demonstration generation process, including the planning, filtering, and code generation, are all performed by LLM. 
\end{itemize}

\section{Related Works}

\subsection{Robot Task Planning}

A common solution to complex robotic tasks is to split them into easier, more manageable sub-tasks. When a robot needs to achieve a goal that cannot be done with a single action, it will have to plan the entire sequence of actions. Most of these approaches solve high-level planning through search-based methods within predefined domains \cite{planning1, planning2, planning3}. Some research also focuses on heuristic strategies for robot task planning \cite{heuristic1, heuristic2, heuristic3}. Recent works leverage LLMs for robot task planning because LLMs can perform multi-task generalization when provided with a designed prompt input. Many recent works utilize prompt engineering for LLMs to generate text as a guide for robot task planning \cite{llmplanning1, llmplanning2, kannan2024smart, liu2024roadmap}. For long horizon tasks, approaches generally plan in a hierarchical manner for the defined task-primitives, and common methods include trees or optimization-based algorithms \cite{kaelbling2011longhorizon, huang2022longhorizon}. In \cite{singh2023progprompt}, the authors introduce situated awareness in LLM-based robot task planning, fulfilling the missing state feedback from the environment when using LLM for task planning. To make the process more deterministic, Liu and Armand \cite{liu2024roadmap} proposed using LLM-generated FSMs. This aligns with our goal of addressing the problem of unseen environmental states. 

\subsection{Imitation Learning for Robotics Manipulation}

In traditional methods, machines and robots could only learn autonomous behaviors when experts spend time writing hard-coded rules for them. Imitation learning offers a way to teach robots by ``showing'' them what to do, simplifying the training of robots to perform new tasks \cite{osa2018immitationlearning1, ravichandar2020immitationlearning2, schaal1999imitationlearning3}. It has demonstrated effectiveness in learning grasping and pick-and-place tasks from low-dimensional states \cite{argall2009ilpickandplace1, billard2004ilpickandplace2}. Some previous works also focus on the few-shot imitation learning, which enables the systems to learn new tasks from just a handful of demonstrations \cite{duan2017ilfewshot1, finn2017ilfewshot2}. 

Language-conditioned imitation learning is an important advancement in this field. It combines natural language instructions with demonstration-based learning to create more flexible and intuitive robot control. This approach allows users to guide robots through natural language commands \cite{jang2022languageconditioned1,brohan2023languageconditioned2, team2024languageconditioned3, cliport}.

Robot demonstration generation is an important concept in the stream of research. LLMs have been used to automatically generate simulations for different purposes \cite{zeng2023llmtaskgen, hu2024agentgen}. Gensim \cite{gensim} introduced an approach that leverages LLMs to automatically generate diverse robotic simulation environments and expert demonstrations. However, this approach has limitations for a wide range of tasks, such as long-horizon tasks or those requiring precise state awareness. Our approach expands the domain of solvable tasks by utilizing LLM-generated finite state machines as guides for demonstration collection.

\section{Method}

\subsection{SMSL Generation via LLM-Based States, Operations and Transitions Filtering} 

The proposed framework uses State Machine Serialization Language (SMSL) to guide demonstration collection for robot manipulation tasks. SMSL is a data language that stores the definition of FSMs in a text format, which can represent the workflow of tasks \cite{liu2024roadmap}. While LLMs could potentially generate plans for complex tasks directly, creating comprehensive planning for complex tasks with a large number of states and transitions from unstructured descriptions is still challenging and lacks scalability. To address this limitation, we implement an indirect SMSL text generation approach that utilizes LLMs to generate and complete executable scripts. The generation process has the following steps:

\begin{itemize}

\item First, the LLM analyzes the task description to list all independent operations (e.g., ``move Object A to Position B'') and find an optimized symbolic representation for the states based on the task requirements (e.g., ``State\_A''). 

\item Second, the LLM defines the state representation and state space, then generates a script that enumerates all possible states through Cartesian product, without considering the task constraints. These states serve as candidates for state filtering.

\item Third, based on the constraints of the task, the LLM constructs filter functions to rule out states that violate the~constraints. After filtering, the LLM will list all valid state-operation pairs and add an operation filter script to prevent transitions from invalid states. It then writes a script to infer state transitions based on the valid state-operation pairs, which will lead to the goal state of this transition. The system applies the state filter again to verify if these goal states are valid; if not, then it removes the entire transition.

\item Finally, the LLM formats the remaining transitions into SMSL text. 

\end{itemize}

The overall algorithm for this LLM-based SMSL generation process is presented in Algorithm \ref{alg:llm_abstract}, with a mathematical formulation of the filter representation for a river-crossing puzzle provided in the Appendix.

\begin{algorithm}[t]
\caption{Indirect SMSL Generation Using LLM}
\label{alg:llm_abstract}
\begin{algorithmic}[1]
\REQUIRE Task description $T$, state constraints $\mathcal{C}_{state}$, operation constraints $\mathcal{C}_{op}$
\ENSURE Valid states, operations, transitions mapping
\\ \textit{Phase 1: Generate Operations and State Candidates}
\STATE $O \leftarrow$ LLM generates operations from $T$
\STATE $S \leftarrow$ LLM generates script to list all states using Cartesian product from $T$ as candidates, ignoring constraints
\\ \textit{Phase 2: Filter States}
\STATE $S_{valid} \leftarrow \{ s \in S \mid$ 
\STATE \hspace{2em} $\text{LLM adds filter that validates } s \text{ against } \mathcal{C}_{state} \}$
\\ \textit{Phase 3: Assign Operations}
\FOR{$s \in S_{valid}$}
    \STATE $O_s \leftarrow$ LLM adds filter that validates operations from $O$ for $s$
\ENDFOR
\\ \textit{Phase 4: Filter Transitions}
\FOR{$(s,o) \in S_{valid} \times O_s$}
    \STATE $s' \leftarrow$ LLM adds code to apply operation $o$ to state $s$
    \IF{$s'$ violates $\mathcal{C}_{state}$ or $o$ violates $\mathcal{C}_{op}$}
        \STATE Remove transition $(s, o)$
    \ENDIF
\ENDFOR
\\ \textit{Phase 5: Generate Output}
\STATE Convert valid transitions to JSON format
\RETURN SMSL in JSON format
\end{algorithmic}
\end{algorithm}

\subsection{State-Aware Task Demonstration Code Design}

Existing approaches like GenSim generate demonstration code for the task that initializes the environments to random object placement \cite{gensim, hua2024gensim2}, which will introduce discrepancies between the task description and the demonstration when applied to longer horizon, more complex tasks. An example of this format of tasks is given in Listing \ref{lst:gensimtask} in Appendix. We design our state-aware task code to establish environment-state persistence via decoupled environment initialization from task logic. Listing \ref{lst:codeformat} in Appendix illustrates what our modified task format looks like. It features three improvements:

\noindent\textbf{Randomize Environment Under Constraints}: In our workflow, a randomized environment will be created for the initial state under the constraints defined in the setup script, guaranteeing compliance with symbolic state requirements.

\noindent\textbf{Environment Passing}: Rather than resetting the environment and spawning the objects in the task code, we will directly pass the environment to the task code.

\noindent\textbf{Deterministic Object Referencing}: To ensure that objects across multiple episodes are referenced deterministically, we create a persistent dictionary which maps each entity (URDF handle) to a runtime object ID.

These changes ensure two functionalities can be performed correctly: First, they free LLMs from locking on low-level objects instantiation towards high-level task goal and constraints understanding. We also expose entity handles to LLMs and allow it to inspect geometric feature when necessary (e.g., chessboard blocks or arbitrary slots), and calculate placement offsets from a target object's pose. Second, it preserves dataset diversity via randomizing all initial pose configurations, while ensuring the state configurations are consistent to the object poses in the environment.

\subsection{Prompt Engineering for Task Generation Agents}

Our code generation pipeline transforms symbolic operations into executable task demonstration scripts by leveraging multiple LLM agents to use our structured prompts. The first stage of the process is broken into two subcomponents, and the second and third stages are recursively applied to each of the operations in the SMSL text:

\noindent\textbf{Agent 0 / Human: Initial-State Environment Setup Script}

An LLM agent will process task descriptions, spatial constraints from expert knowledge and scene entity library references to produce the initialization scripts that: 

\begin{itemize}
    \item Spawn objects at random poses under geometric constraints \textit{only} for the initial state.
    \item Establish persistent mappings between scene entities and simulation object IDs.
\end{itemize}

Verifications can be made to ensure this initialization script can have the spawned environment consistent with the initial state's configurations before passing it to the next LLM agent.

\noindent\textbf{Agent 1: Task Specification Generation}

Another LLM agent will read through this initial state environment setup script and task description to generate a structured task specification in JSON format containing:

\begin{itemize}
    \item Entity List: Categorized URDF file names of used meshes by their physics types (fixed or rigid).
    \item Entity to Inspect: A URDF file name that requires geometric inspection by the task generation agent.
    \item Natural language task summary.
\end{itemize}

\noindent\textbf{Agent 2: Task Code Generation}

The final LLM agent will read through CLIPort's API to understand the definition of the ``task'' class, then to review examples of success and failure with the type of failure. It creates the task code according to the JSON description extracted from the task specification and also inspects the URDF if it's listed in ``entity to inspect'', so that it knows how to properly set the offset for the target pose.

\subsection{Demonstration Dataset Generation}

\begin{figure}[t]
    \centering
    \includegraphics[width=\columnwidth]{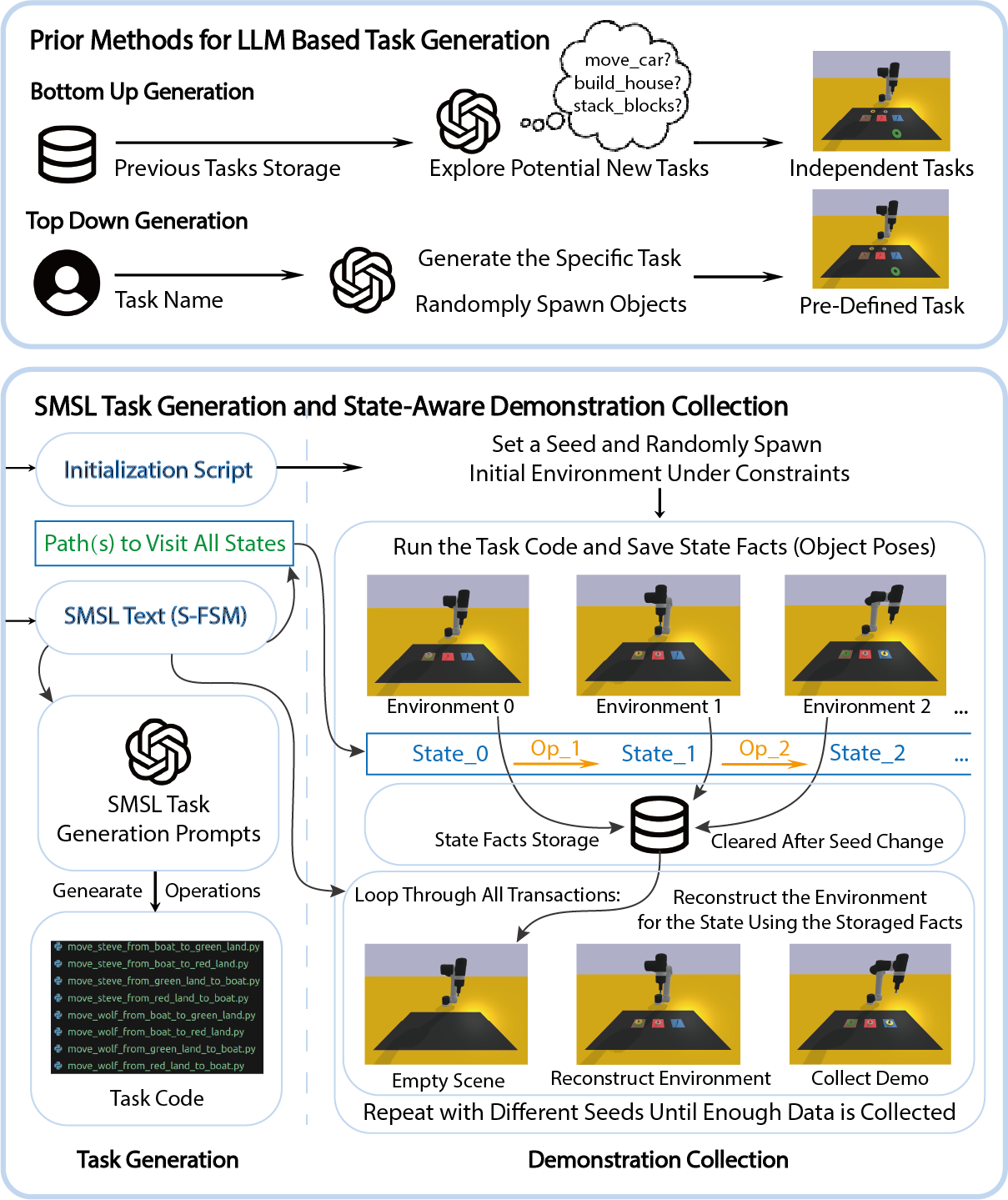}
    \caption{Comparison of demonstration generation workflows. The existing bottom-up approach is to let LLM to explore potential new tasks from the previous task list. The existing top-down approach will use the user-defined task name and generate code for this specific task. Both of these methods will reset the environment and spawn objects randomly on the tabletop when generating a demo, which are not suitable for long-horizon tasks or tasks with dynamically evolving spatial constraints.}
    \label{fig:compare_workflow}    
\end{figure}

Our framework provides automatic demonstration dataset generation in a two-phase approach using the constructed FSM, the workflow and the comparison with the existing LLM-centralized workflow for task generation is illustrated in Fig. \ref{fig:compare_workflow}. At first, we apply graph search approach in order to find either a single path that covers all the states or, if that fails, generate multiple minimum-length paths to cover all states: We employ a depth-first search that maximizes state coverage; the algorithm seeks to find a single path containing every state by tracking visited states with bitmasking and if failed, produces several minimum-length paths using a coverage-optimized queue. These paths provide guidance for exploration of states, which also infers the sequence of operations applied to the initial environment. 

The second stage uses the paths to generate training demonstrations for CLIPort \cite{cliport}. It starts with spawning an environment of the initial state and storing the accurate state configurations in the storage. Then, it applies the first operation in the path to the environment and collects the state configurations as well. This process is then repeated until it has collected all or exemplary state configurations in case of exceedingly large state space. Following this, the framework loops over all of the transitions as defined in the SMSL, reconstructs the environment from the stored configurations, performs the operation, and generates the demonstration. To be more specific, we denote the object poses for each transition $(s_i,o_j\rightarrow s_k)$ in the path as:

\begin{equation}
    F_{s_i} = \{e \mapsto \text{get\_object\_pose}(id_e) \mid e \in E\}
\end{equation}

where $E$ represents the set of all entities (objects) in the environment, ``get\_object\_pose'' is a script function that will take the object id as input and output the object pose. During each transition $(s_i, o_j \rightarrow s_k)$, the environment is initialized with constraint-compliant random poses, followed by saving the object poses as state configurations $F_{s_i}$. The system then executes operation $o_j$ using LLM-generated task scripts with the current environment and save the resulting state configurations as $F_{s_k}$. This process continues along the paths until all state configurations are collected, where these configurations ${F_s}$ are stored in a state dictionary unique to each randomized initialization. This approach enables deterministic state reconstruction and effectively separates the demonstration collection process from environment randomization.

The demonstration collection iterates through all transitions in SMSL (in any order, allowing parallel execution), where for each transition, the system spawns an environment using stored state configurations, executes the corresponding task code, and collects the required data for CLIPort (RGB-D observations, language goals, and actions). The environment is then initialized in random poses that are consistent with constraints.

\subsection{Design of Experiments}

The experimental evaluation includes three complex tabletop manipulation tasks, each representing a distinct finite state machine (FSM) with varying complexity levels according to Table \ref{tab:fsmconfig}. 

\noindent\textbf{Towers of Hanoi:} The set-up includes three stands (blue, red, and brown) and three rings: gray, yellow, and green in increasing radius. Following standard rules, only smaller rings can be placed on larger ones, the robot can only move one of the rings at a time. The three stands are spawned randomly on the table while the rings are all located on the brown stand.

\noindent\textbf{River Crossing:} A puzzle with the Human, Wolf, Sheep, and Grass entities that initially begin in four on red land. However, the sheep cannot be alone with both the wolf and grass in the absence of Human. For any two-object boat transport, Human must be present. A detailed task description and constraints representation are provided in the Appendix.

\begin{figure}[t]
    \centering
    \begin{subfigure}[b]{0.3\columnwidth}
        \includegraphics[width=\textwidth]{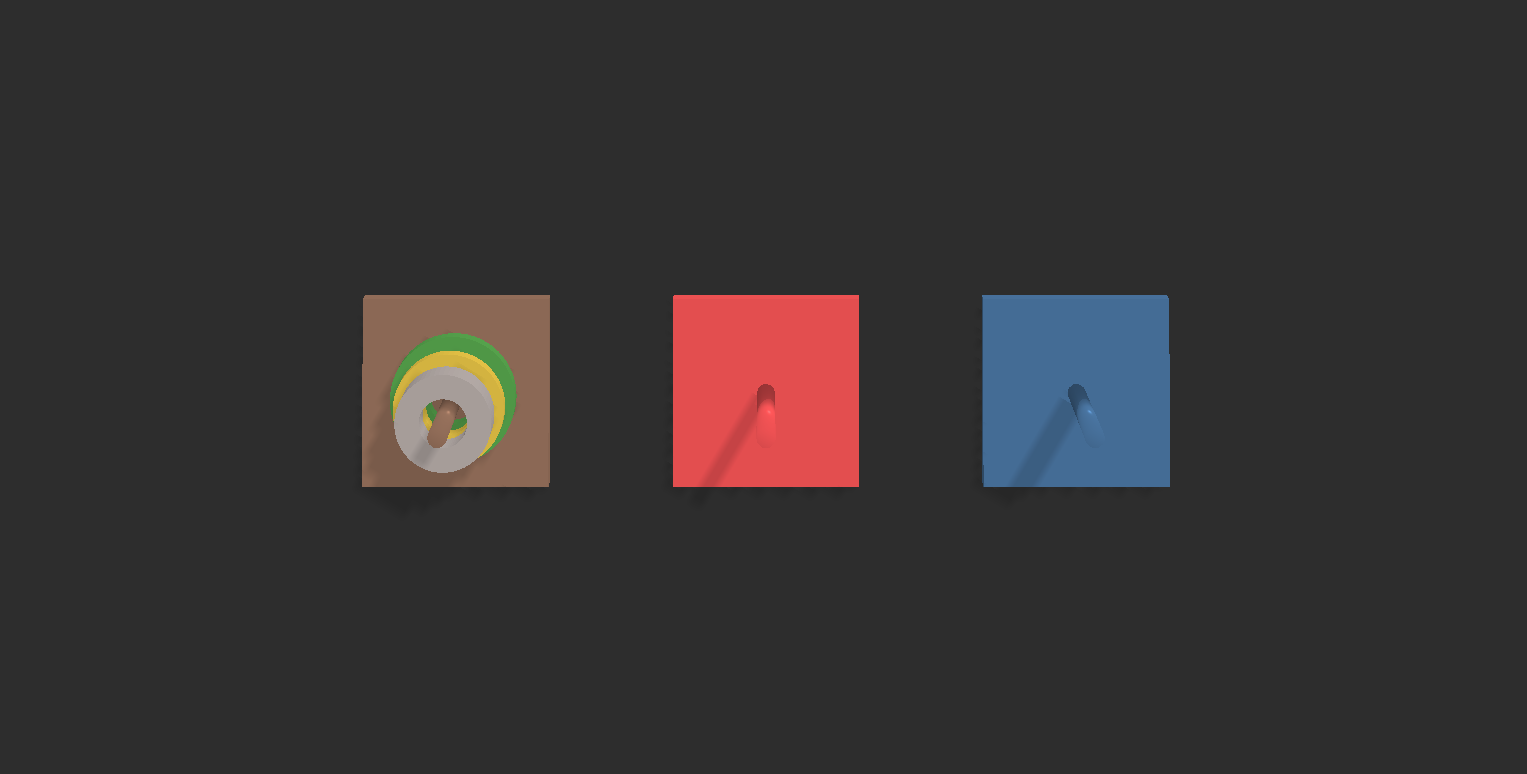}
        \caption{Hanoi}
        \label{fig:hanoi}
    \end{subfigure}
    \hfill
    \begin{subfigure}[b]{0.3\columnwidth}
        \includegraphics[width=\textwidth]{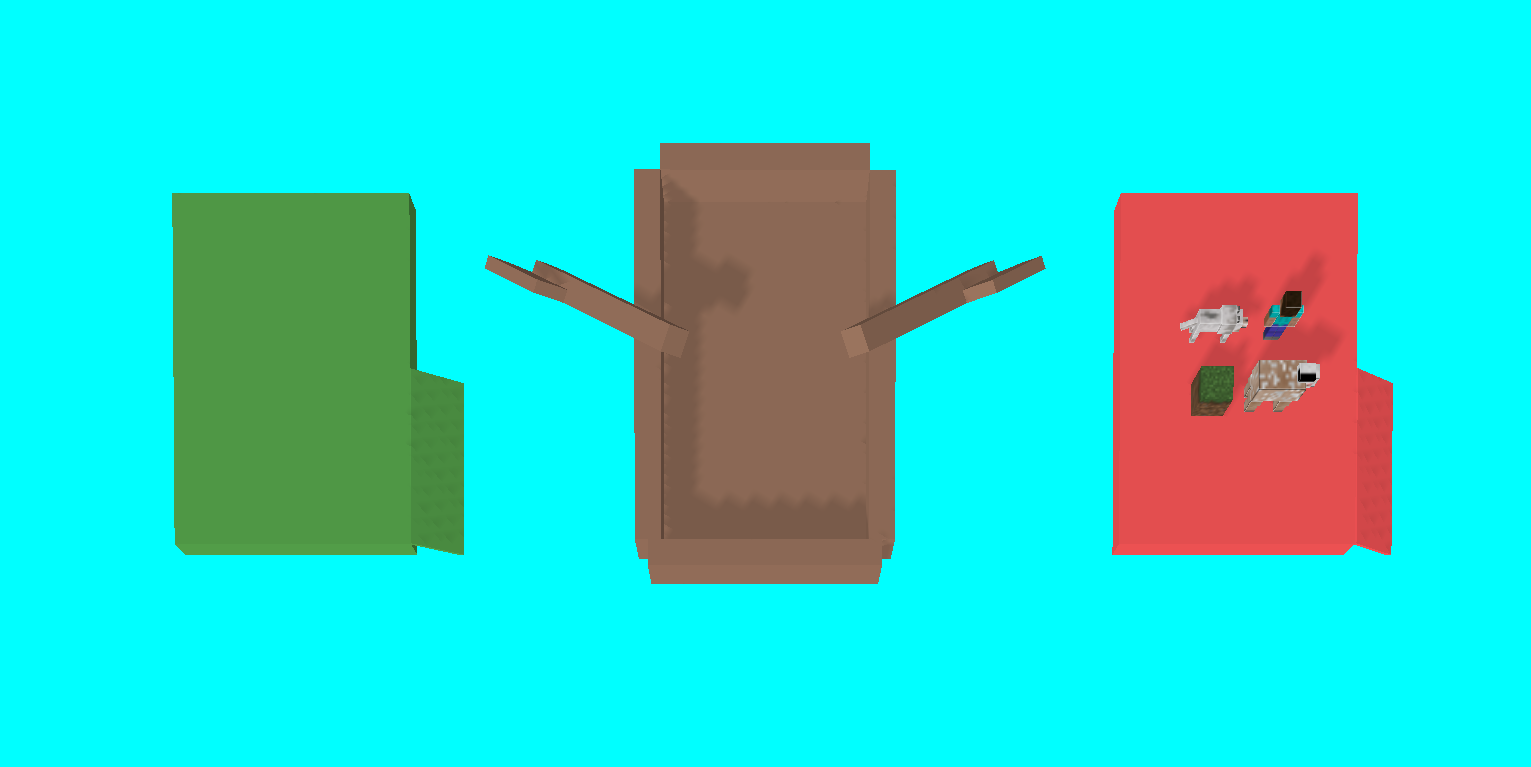}
        \caption{River Crossing}
        \label{fig:river}
    \end{subfigure}
    \hfill
    \begin{subfigure}[b]{0.3\columnwidth}
        \includegraphics[width=\textwidth]{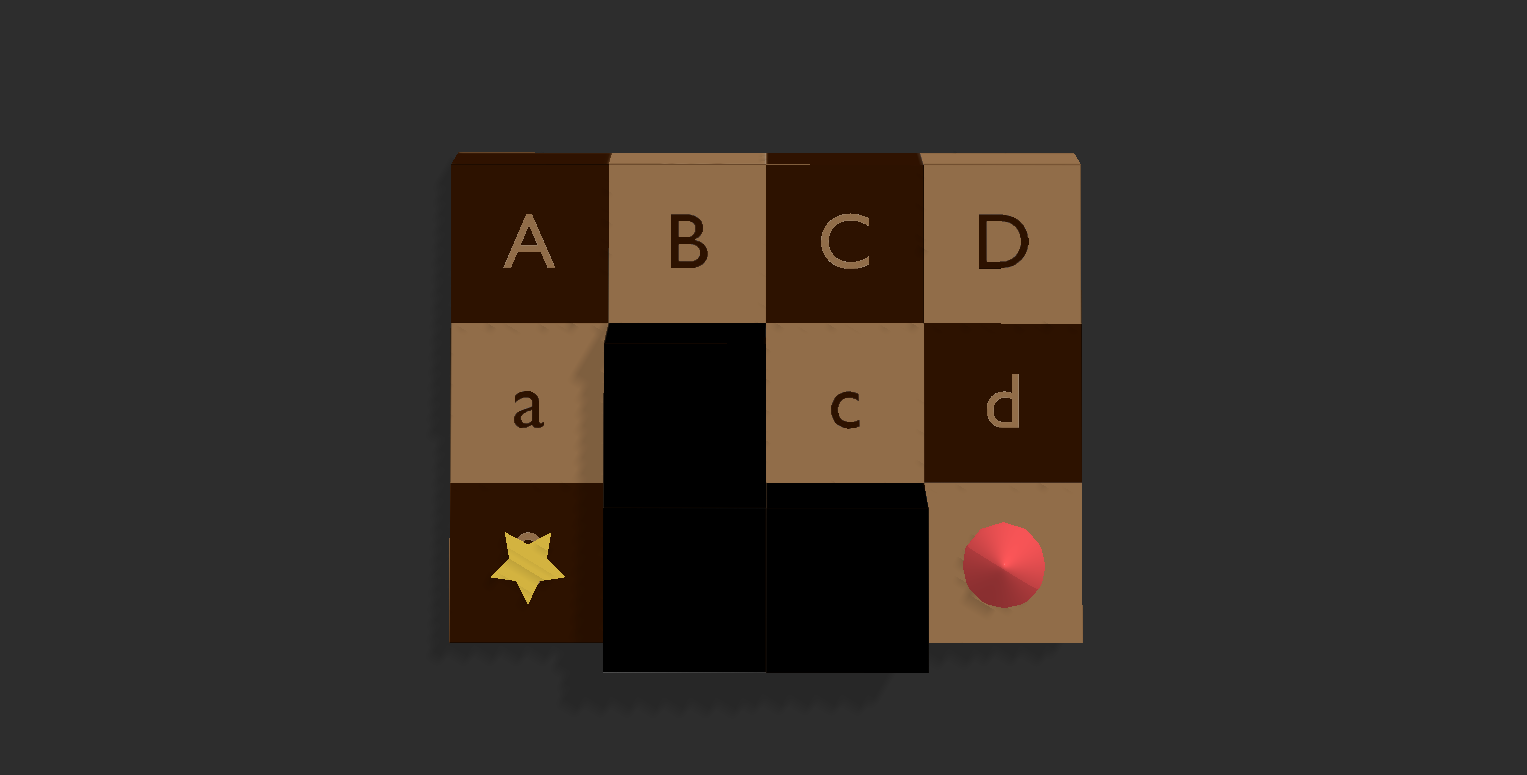}
        \caption{Chess}
        \label{fig:chess}
    \end{subfigure}
    \caption{Task initial scene}
    \label{fig:tasks}
\end{figure}

\noindent\textbf{Chess:} The Chess Board puzzle has nine blocks and two chess pieces. Among the nine blocks, three of them are impassable. Pieces can be placed on top of each other in the same block, and the state ``star chess is on top of the circle chess'' is different than the state ``circle chess is on top of the star chess''. At the start of the game, the star chess is spawned at the bottom left block as seen in Fig. \ref{fig:chess}, and the circle chess is spawned at the bottom right block. The goal is to switch the positions of the pieces.

\begin{table}[h]
\caption{Task Finite State Machine Configurations}
\label{tab:fsmconfig}
\begin{center}
    \begin{tabular}{lccc}
        \toprule
        & \textbf{Towers of Hanoi} & \textbf{River Crossing} & \textbf{Chess} \\
        \midrule
        Rigid Objects & 3 & 4 & 2 \\
        Valid Operations & 9 & 16 & 18 \\
        Valid States & 27 & 40 & 90 \\
        Valid Transitions & 78 & 92 & 324 \\
        \bottomrule
    \end{tabular}
\end{center}
\end{table}

\section{Results and Discussion}

Notably, unlike traditional puzzle-solving scenarios, our implementation does not focus on reaching specific goal states. Instead we use these designed puzzle tasks to illustrate complex FSM navigation, where intermediate states hold equal importance to final states. It better represents the real world applications, where there could be many transitions in states and intermediate configurations that are equally valid targets. By applying this philosophy to our sample collection process, we collect demonstrations covering the entire state space or exemplary states if the state space is too large, rather than only goal-oriented trajectories.

We evaluated our method by training on our collected dataset across three puzzles. We trained the model with 200, 500, and 1000 demonstrations for each operation, and we used the average success rate across all operations as the evaluation metric. To highlight the necessity of state-aware demonstration collection for these complex long-horizon tasks, we compared our method against a control condition where all objects are randomly placed on the table when collecting demos - a typical setting in previous methods. We also trained with 200, 500, and 100 demonstrations per operation for this control condition and tested in the same puzzle-solving contexts. Results demonstrate that the policy trained on random object placement demos does not generalize well to situations where objects need to understand dynamically changing state dependencies. Table \ref{tab:results} presents the detailed comparative results.

\begin{table}[htbp]
\centering
\caption{Performance Comparison Across Tasks and Demo Numbers.}
\label{tab:results}
\label{tab:results}
\begin{tabular*}{\columnwidth}{@{}p{2.8cm}@{\extracolsep{\fill}}ccc@{}}
\toprule
\textbf{Method} & \multicolumn{3}{c}{\textbf{Number of Demonstrations}} \\
\cmidrule(lr){2-4}
 & \makebox[0.2\columnwidth]{200} & \makebox[0.2\columnwidth]{500} & \makebox[0.2\columnwidth]{1000} \\
\midrule
\multicolumn{4}{@{}l@{}}{\textbf{Hanoi} (tolerance 0.01m, 15 degrees)} \\
Our Method & 0.849 & 0.942 & 0.980 \\
GenSim & 0.587 & 0.606 & 0.580 \\
\midrule
\multicolumn{4}{@{}l@{}}{\textbf{River Crossing} (tolerance 0.06m, 15 degrees)} \\
Our Method & 0.716 & 0.947 & 0.962 \\
GenSim & 0.027 & 0.077 & 0.055 \\
\midrule
\multicolumn{4}{@{}l@{}}{\textbf{Chess} (tolerance 0.01m, 15 degrees)} \\
Our Method & 0.890 & 0.929 & 0.980 \\
GenSim & 0.394 & 0.443 & 0.430 \\
\bottomrule
\end{tabular*}
\end{table}

We show that with LLM-based simulation generation, our proposed method can successfully perform long-horizon complex tasks while the control condition (randomly placing objects on the table) fails. Our contributions allow LLM-powered robots to be capable of performing real world tasks where environmental states must be rigorously tracked and propagated across long-horizon complex tasks.

\section{Conclusion}

Current demonstration generation approaches using LLMs in imitation learning frameworks are limited by the coverage of the sample set. When the demonstration set lacks examples for dynamically evolving situations, the model may fail to generalize, leading to cascading errors during robotic execution. Our experiments confirm that existing frameworks are particularly prone to such failures in long-horizon tasks. To address this limitation, we propose a state-aware approach that enhances demonstration coverage. Our implementation improves the success rate of long-horizon tasks, achieves an up to 98\% success rate in these tasks, compared to the controlled condition for the existing approach, which only had a success rate of up to 60\%, and some tasks experience catastrophic failure.

\newpage

\section*{Appendix}

\subsection{Formalism of the River Crossing Puzzle}

To demonstrate our methodology, here we use the river crossing task as an example, which is a task with constraints that requires reasoning while robot moving. The game involves four entities—Human, Sheep, Wolf, and Grass. They must be transported from red land to green land via a boat with limited capacity. We define the following constraints to the task: 

\begin{itemize}
    \item The robot can move one entity per operation between adjacent locations (red land - boat - green land).
    \item Sheep cannot coexist with Wolf or Grass without the Human presence.
    \item The boat holds no more than two entities, requiring Human's presence when occupied by two.
\end{itemize}

We now present our LLM-based symbolic planning approach, through the following formalization steps.

\subsubsection{State Space Definition}

We formalize the problem as a 4-tuple $T=(P,O,A,C)$ where:
\begin{itemize}
    \item 1. $P=\{\text{Red\_land}, \text{Boat}, \text{Green\_land}\}$ (Positions)
    \item 2. $E=\{\text{Human}, \text{Sheep}, \text{Wolf}, \text{Grass}\}$ (Entities)
    \item 3. $O$: 16 possible move operations between positions
    \item 4. $C$: Safety and capacity constraints
\end{itemize}

The complete state space is generated through a Cartesian product:
\begin{equation}
    \mathcal{S}_{\text{candidate}} = \prod_{l \in L} S = 3^4 = 81~\text{states}
\end{equation}

Each state is encoded as:
\begin{equation}
    s = (s_{\texttt{Human}}, s_{\texttt{sheep}}, s_{\texttt{wolf}}, s_{\texttt{grass}}) \in S^4
\end{equation}

\subsubsection{States Filtering} 

Here for this river-crossing task, we define the constraint as $C = C_{\text{safety}} \wedge C_{\text{capacity}}$:

\begin{equation}
\begin{aligned}
\mathcal{C}_{\text{safety}}: \forall e &\in E \setminus \{\text{human}\}, \\
(s_e = s_{\text{sheep}} &\land e \in \{\text{wolf}, \text{grass}\}) &\rightarrow s_{\text{human}} = s_{\text{sheep}}
\end{aligned}
\end{equation}

\begin{equation}
\begin{aligned}
\mathcal{C}_{\text{capacity}}: 
&\sum_{e \in E} \mathbb{I}(s_e = \text{boat}) \leq 2 \\
&\land \bigl(\sum_{e \in E} \mathbb{I}(s_e = \text{boat}) = 2 &\phantom{\land} \rightarrow s_{\text{human}} = \text{boat}\bigr)
\end{aligned}
\end{equation}

\noindent where $\mathbb{I}$ is the indicator function that equals 1 when the condition is true and 0 otherwise, $\mathcal{C}_{\text{safety}}$ corresponds to the second constrains and $\mathcal{C}_{\text{capacity}}$ corresponds to the third constraint. For example, state 
$({\text{boat}}, {\text{green}}, {\text{boat}}, {\text{red}})$ violates 
$\mathcal{C}_{\text{safety}}$ (sheep and wolf unsupervised), while 
$({\text{red}}, {\text{boat}}, {\text{boat}}, {\text{boat}})$ violates 
$\mathcal{C}_{\text{capacity}}$ (3 entities in boat). The LLM agent will generate filter code that implements the above logic. After applying these constraints, the state space is reduced from 81 state candidates to 40 valid states.

\subsubsection{Operations and Transitions Filtering} 

We here define a transition to be the complete path of a source state, an operation, and a resultant state. Valid transitions require operation-specific preconditions and post-validation. For a transition $t = (e, s_{\text{src}}, s_{\text{dest}})$:

\begin{equation}
\begin{aligned}
\text{Pre}(o,s) = & (s_e = s_{\text{src}}) \ \land \\
& \left((s_{\text{src}} = \text{boat} \land s_{\text{dest}} \neq \text{boat}) \right. \\
& \left. \lor (s_{\text{src}} \neq \text{boat} \land s_{\text{dest}} = \text{boat} \right. \\
& \left. \land \mathcal{C}_{\text{capacity}}(s'))\right)
\end{aligned}
\end{equation}

\noindent where $\text{Pre}(o,s)$ is the precondition for operation $o$ in state $s$, $s_e$ is the location of entity $e$, $s_{\text{src}}$ is the source location, $s_{\text{dest}}$ is the destination location, and $\mathcal{C}_{\text{capacity}}(s')$ is a constraint ensuring the boat capacity isn't exceeded in the resulting state $s'$. This precondition states that an entity must be at the source location, and either we are moving from the boat to a land, or from a land to the boat (if capacity allows). For example, moving sheep from boat to green land requires: 1) Sheep is in boat, 2) Human remains with Wolf/Grass, and 3) Boat capacity remains valid. LLM agent generate executable filter script for the same logic, and after applying these LLM-generated filters, we can construct an accurate SMSL text for the FSM with much less human validation and correction compared to direct generation using LLM.

\subsection{Task Demonstration Generation Code}

\begin{lstlisting}[caption={GenSim Task Demonstration Code Format},label={lst:gensimtask}]
# env is initialized with no objects
super().reset(env)
...
obj_pose = self.get_random_pose(env, size0)
obj_id = env.add_object(urdf, obj_pose, ...)
targ_pose = self.get_random_pose(env, size1)
targ_id = env.add_object(urdf, targ_pose, ...)
...
self.add_goal(objs=[obj_id], targ_poses=[targ_pose], ...)
\end{lstlisting}

\begin{lstlisting}[caption={Our Task Demonstration Code Format},label={lst:codeformat}]
# env is passed into the function
obj_id = env.asset_ids_dict['rigid']["x.urdf"]
targ_id = env.asset_ids_dict['fixed']["y.urdf"]
targ_pose = env.get_object_pose(targ_id)
self.add_goal(objs=[obj_id], targ_poses=[targ_pose], ...)
\end{lstlisting}

\newpage

\bibliographystyle{IEEEtran}
\bibliography{references}

\end{document}